\PassOptionsToPackage{unicode}{hyperref}
\PassOptionsToPackage{hyphens}{url}
\documentclass[
  11pt,
]{article}
\usepackage{amsmath,amssymb}
\usepackage{iftex}
\ifPDFTeX
  \usepackage[T1]{fontenc}
  \usepackage[utf8]{inputenc}
  \usepackage{textcomp} 
\else 
  \usepackage{unicode-math} 
  \defaultfontfeatures{Scale=MatchLowercase}
  \defaultfontfeatures[\rmfamily]{Ligatures=TeX,Scale=1}
\fi
\usepackage{lmodern}
\ifPDFTeX\else
\fi
\IfFileExists{upquote.sty}{\usepackage{upquote}}{}
\IfFileExists{microtype.sty}{
  \usepackage[]{microtype}
  \UseMicrotypeSet[protrusion]{basicmath} 
}{}
\makeatletter
\@ifundefined{KOMAClassName}{
  \IfFileExists{parskip.sty}{%
    \usepackage{parskip}
  }{
    \setlength{\parindent}{0pt}
    \setlength{\parskip}{6pt plus 2pt minus 1pt}}
}{
  \KOMAoptions{parskip=half}}
\makeatother
\usepackage{xcolor}
\usepackage[margin=1in]{geometry}
\usepackage{color}
\usepackage{fancyvrb}

\DefineVerbatimEnvironment{Highlighting}{Verbatim}{commandchars=\\\{\}}
\newenvironment{Shaded}{}{}

\newcommand{\AttributeTok}[1]{\textcolor[rgb]{0.49,0.56,0.16}{#1}}

\newcommand{\BuiltInTok}[1]{\textcolor[rgb]{0.00,0.50,0.00}{#1}}

\newcommand{\CommentTok}[1]{\textcolor[rgb]{0.38,0.63,0.69}{\textit{#1}}}

\newcommand{\ControlFlowTok}[1]{\textcolor[rgb]{0.00,0.44,0.13}{\textbf{#1}}}
\newcommand{\DataTypeTok}[1]{\textcolor[rgb]{0.56,0.13,0.00}{#1}}
\newcommand{\DecValTok}[1]{\textcolor[rgb]{0.25,0.63,0.44}{#1}}

\newcommand{\FunctionTok}[1]{\textcolor[rgb]{0.02,0.16,0.49}{#1}}

\newcommand{\KeywordTok}[1]{\textcolor[rgb]{0.00,0.44,0.13}{\textbf{#1}}}
\newcommand{\NormalTok}[1]{#1}
\newcommand{\OperatorTok}[1]{\textcolor[rgb]{0.40,0.40,0.40}{#1}}

\newcommand{\SpecialCharTok}[1]{\textcolor[rgb]{0.25,0.44,0.63}{#1}}

\newcommand{\StringTok}[1]{\textcolor[rgb]{0.25,0.44,0.63}{#1}}

\newcommand{\VerbatimStringTok}[1]{\textcolor[rgb]{0.25,0.44,0.63}{#1}}

\usepackage{graphicx}
\makeatletter
\newsavebox\pandoc@box
\newcommand*\pandocbounded[1]{
  \sbox\pandoc@box{#1}%
  \Gscale@div\@tempa{\textheight}{\dimexpr\ht\pandoc@box+\dp\pandoc@box\relax}%
  \Gscale@div\@tempb{\linewidth}{\wd\pandoc@box}%
  \ifdim\@tempb\p@<\@tempa\p@\let\@tempa\@tempb\fi
  \ifdim\@tempa\p@<\p@\scalebox{\@tempa}{\usebox\pandoc@box}%
  \else\usebox{\pandoc@box}%
  \fi%
}
\def\fps@figure{htbp}
\makeatother
\providecommand{\tightlist}{%
  \setlength{\itemsep}{0pt}\setlength{\parskip}{0pt}}
\usepackage{fvextra}
\DefineVerbatimEnvironment{Highlighting}{Verbatim}{commandchars=\\\{\},breaklines,breakanywhere,fontsize=\small}
\usepackage{bookmark}
\IfFileExists{xurl.sty}{\usepackage{xurl}}{} 
\hypersetup{
  pdftitle={Applying Anthropic Primitives at Large Enterprises: Harness Paradigm for Knowledge Work},
  pdfauthor={George Salapa, G.S. s.r.o., PwC Austria},
  hidelinks,
  pdfcreator={LaTeX via pandoc}}

\title{Applying Anthropic Primitives at Large Enterprises: Harness
Paradigm for Knowledge Work}
\author{George Salapa, G.S. s.r.o., PwC Austria}
\date{}

\begin{document}
\maketitle

\section*{Abstract}\label{abstract}
\addcontentsline{toc}{section}{Abstract}

Frontier models have collapsed the cost of writing custom code: a niche
problem a specialist sees inside their own domain, too small to ever
earn central budget, now costs an afternoon. The cost of reviewing,
understanding, and maintaining that code afterwards hasn't collapsed at
all. Each specialist's solution drifts from the next, and understanding
one means reading its codebase from scratch. That cost is real enough
that this practice goes largely unendorsed in an enterprise setting, and
management still, reasonably, wants a transparent view of what's
running. Large enterprises build something centrally provisioned and
governed instead: at worst an off-the-shelf product, at best a
graph-orchestration framework such as LangChain wired bespoke per use
case, or a low-code conversational platform such as Copilot Studio used
as the orchestrator itself. These patterns are custom every time and
limited in what they can do. The harness paradigm, gaining ground in
recent months, is neither.

A growing body of recent work treats the coding-agent harness as
enterprise infrastructure rather than a coding tool. It converges on
three findings: harnesses suffice at the task level and outperform more
elaborate agent architectures on enterprise work (arXiv:2604.00073,
arXiv:2604.13107); harness choice accounts for most of the variance in
agent benchmark results, more than model choice does (arXiv:2605.23950);
and the open gap standing between that finding and enterprise adoption
is governance (arXiv:2605.10223, arXiv:2605.18747).

Informed by engagements across several enterprise clients, we propose an
architecture that aims to close that gap. One harness runs unmodified as
the backbone, the underlying code stays identical across every
deployment. Reviewing what gets built thus collapses to reading its
instructions file. Today enterprises keep the harness at the engineer's
terminal and build the actual orchestration itself for every new use
case, wiring a graph-orchestration framework bespoke, standing up a
low-code platform as the orchestrator, or bolting a frontier model onto
existing software as a glorified autocomplete. We argue for the
opposite: the harness itself belongs in that spot.

Section 4 gives four mechanisms that aim to achieve this. First,
credential-scoped tooling. We don't build a separate method for every
operation a tool might perform. Each backend, SharePoint, a SQL
warehouse, whatever, gets one generic request tool and a scoped,
identity-bound credential instead. The gateway decides which backends
you're allowed to reach at all; once you're in, the model composes
whatever call it needs itself. This is what lets governance survive
contact with the enterprise's actual identity and access system: access
is controlled at the credential, not policed operation by operation.
Second, authorization logic never lives inside the harness itself, so
one harness artifact runs unmodified as an unattended cron backbone, as
the execution engine behind a business-facing chat surface, and
interactively at a terminal, all under a single identity and governance
model. Third, every deployment registers itself as a side effect of
shipping, so auditing an enterprise's N agent solutions collapses to
reviewing N version-controlled instruction files. Fourth, a call the
model flags risky is reviewed by a freshly spawned instance of the same
harness before any human sees it, a fresh context judging the call
instead of the one that composed it grading its own work.

All four rest on the same idea: an engineer forks the reference
repository, edits an instructions file and a configuration file, and
pushes. CI/CD builds a container, registers the solution, and deploys
it, wired to a tool registry (what it may call), a solution registry
(what was deployed, by whom), and a skill library (procedural knowledge
baked in at build time by default, or fetched live from GitHub for any
skill named live in configuration). The only other moving part is a
run-trigger endpoint, which lets a chat surface such as Copilot Studio
launch that same container on demand.

We built the architecture described here on microcc
(\url{https://pypi.org/project/micro-cc/}), our reference harness
implementation used throughout this paper.

\section*{1. Introduction}\label{introduction}
\addcontentsline{toc}{section}{1. Introduction}

Enterprises today run on a landscape built from four familiar patterns.
One team builds a retrieval-augmented pipeline on a framework such as
LangChain or LlamaIndex. Another wires a different graph for a different
problem on Semantic Kernel, CrewAI, or an equivalent amount of custom
Python; it's its own pipeline end to end, sharing no tool layer or
governance model with the first. A third creates a chatbot for business
users on a low-code conversational platform such as Copilot Studio, and
uses it as the orchestrator itself. A fourth deploys an internal chatbot
on a frontier model with web search and a limited, no-file-access code
interpreter: it reasons at a state-of-the-art level but can't open a
file on the company's own document store. None of these four shares a
codebase, a tool registry, or a governance model with any of the others.
Every new use case an enterprise takes on starts the same integration
work again from scratch.

Several patterns emerge across all four. The model sits as a layer
bolted onto existing software: a glorified autocomplete summoned to
answer one siloed slice of a problem. Tools are engineer-defined; a
catalog of hand-built methods, one per anticipated action. That catalog
grows linearly with every new system and operation a team thinks to
anticipate. Architecture gets chosen per surface, too. A chatbot team
builds on a low-code conversational platform, an automation team wires a
graph-orchestration framework, a developer-tools team adopts a terminal
harness, and each works independently, on its own codebase and library
of notes. Governance is permission-seeking: before a team can write a
line of code, it asks a review board or security function whether it's
allowed to build at all, and waits. A lot of valuable innovation moves
off-radar as a result.

Another pattern: enterprises build their logic directly on a low-code
conversational platform such as Copilot Studio, using it as the
orchestrator itself. Governance and audit come largely for free,
procurement is already signed, and a business team can ship without
engineering support, which is a real advantage. It comes at a cost,
though. These platforms are weak orchestrators, failing to reason across
more than a handful of connected actions, and their internal decision
path is a black box the team that owns the outcome can't inspect or
change. Section 4.9 argues for a narrower role for platforms of this
kind: thin ingress and identity.

A separate body of recent work treats the coding-agent harness as
infrastructure rather than a coding tool. It converges on three
findings. Harnesses suffice at the task level and outperform more
elaborate agent architectures on enterprise work (arXiv:2604.00073,
arXiv:2604.13107). Harness choice now accounts for most of the variance
in agent benchmark results, more than model choice does
(arXiv:2605.23950). And the open gap standing between that finding and
enterprise adoption is governability (arXiv:2605.10223,
arXiv:2605.18747). None of this work proposes the architecture that
closes that gap. Closing it means running a harness safely behind an
enterprise's identity system and audit trail, at the scale of many teams
building in parallel, and that's what this paper does.

Section 4 proposes this architecture. Giving a model hands, a loop,
memory, filesystem access, is a new basic unit of software. The
API-first decade made software legible to other software. The next
frontier is software legible to a model, and the fundamental building
block becomes the model in a box: a loop, memory, filesystem, bash.
Boxes compose via message passing (Section 4.5), and the engineering
effort moves to the walls between them.

The harness is the backbone, unmodified. An engineer forks it, writes an
instructions file and a configuration file, and pushes code. CI/CD
builds a container and drops it into deployment. Four thin services hold
the fleet together: a tool registry gating what any running container
may call, a solution registry recording what was deployed and by whom, a
skill library it draws procedural knowledge from, baked into the image
at build time by default or fetched live at runtime for any skill a
configuration names live, and a run-trigger endpoint that lets a chat
surface such as Copilot Studio or Slack launch that same container on
demand. Section 4 explains this in detail.

We propose an alternative to each of these patterns.

Enterprises today curate tools per operation: one hand-built method for
each action someone anticipated. We build the tool catalog around the
credential instead. Each backend gets one generic request tool and a
scoped, identity-bound token; the model composes whatever call it needs
against that token itself. We call this credential-scoped tooling
(Section 4.5). Because access is decided at the credential rather than
looked up per operation, governance survives contact with the
enterprise's real identity and access system.

Governance is a natural consequence of deployment. Every deployment
registers itself as a side effect of shipping, so auditing an
enterprise's N agent solutions collapses to reviewing N
version-controlled instruction files (Sections 4.7, 4.8, 4.11).

\section*{2. Background / Related Work}\label{background-related-work}
\addcontentsline{toc}{section}{2. Background / Related Work}

The architecture described in this paper is informed by work in large
enterprises in Europe, spanning automotive, manufacturing, fast-moving
consumer goods, and healthcare. Individual client or employer
engagements are not named or described here; the claims made are about
the architecture, and should be evaluated on that basis. The mechanisms
below were proposed and evaluated against an Azure cloud environment:
its directory groups, management groups, and role-based access control
shaped how they were implemented, not their underlying logic. The same
architecture is directly portable to other cloud providers, and most of
its mechanisms can be lifted into an on-premises setting as well.

Recent work has begun to treat the coding-agent harness as
infrastructure rather than a coding tool.

Terminal Agents Suffice for Enterprise Automation (arXiv:2604.00073)
argues that a terminal-and-filesystem agent matches or outperforms more
complex MCP- or GUI-based agent architectures on enterprise tasks, and
that simple programmatic interfaces combined with strong foundation
models could be the backbone of enterprise automation. Section 4 shares
that claim.

Can Coding Agents be General Agents? (arXiv:2604.13107) tests a coding
agent against an open-core ERP system and finds that simple tasks
succeed reliably even with no ERP-specific tooling at all, which rules
out tool access as the bottleneck and places it instead in bridging
domain logic and code execution as task complexity rises. Four failures
occur: lazy heuristics (a stated policy implemented with a crude
shortcut, such as filtering vendors by name instead of an address
field), hallucination (the agent inventing and then acting on a
nonexistent system state), dropped constraints (an explicit rule simply
not carried forward through a long reasoning trace), and overconfidence
(the agent reporting success regardless of outcome, because code-level
feedback is dense while business-level feedback is thin). The tool
gateway in Section 4.2 answers a narrower question than any of these:
whether the agent can safely reach a system at all. It doesn't resolve
any of the four failure modes itself. The deployment path in Section
4.7, running the task interactively first and distilling what worked
into the instructions file, mitigates them in practice; that matches our
experience building this way across engagements.

Code as Agent Harness (arXiv:2605.18747) surveys code as the operational
substrate of agent systems across coding, GUI/OS automation, and
enterprise workflows, and names open challenges including consistent
shared state across multiple agents and human oversight for
safety-critical actions. Sections 4.5 and 4.6 address both directly. A
model has no continuity of its own between calls; what it retains across
runs is the externally engineered state Section 4.5 describes. And the
mechanism Section 4.6 shows in code, a blocking approval queue that
gates a high-risk tool call the moment it's made rather than after the
fact, is the answer to human oversight.

Dive into Claude Code: The Design Space (arXiv:2604.14228) compares
Claude Code against two independent systems, OpenClaw and Hermes Agent,
and finds each answers the same recurring design questions differently,
occupying a different region of the design space depending on deployment
context, though the paper treats the three as composable pieces of one
larger design space. Section 4 takes that same comparison further: one
harness core supplies all three deployment contexts itself, under a
single identity and governance model.

Beyond Autonomy: A Dynamic Tiered AgentRunner Framework
(arXiv:2605.10223) argues that current agent frameworks prioritize
autonomy over the governability enterprise deployment requires, and
proposes risk-tiered review; the entitlement and approval mechanisms in
Sections 4.6 to 4.8 are one concrete instance of that argument. Their
proposal is considerably more elaborate than what is described in this
paper. Section 4 offers something narrower: a small set of operational
primitives (identity-scoped entitlement, a registration gate, a blocking
approval queue) that extend the existing single-loop harness from a
local developer tool into an enterprise backbone. There's no separate
governance protocol and no second population of agents to run it.

Stop Comparing LLM Agents Without Disclosing the Harness
(arXiv:2605.23950) establishes that harness choice now accounts for most
of the variance in agent benchmark results, more than model choice does.
This paper does not run a comparative benchmark between harnesses at
all, and in the spirit of that finding discloses its own directly: the
work described throughout was built on microcc
(\url{https://pypi.org/project/micro-cc/}), used here as a reference
implementation.

Distributing Security Controls Through Harness Engineering (SHarD,
arXiv:2607.25890) reaches this paper's thesis from a security angle
rather than a governance one. It lands in the same place: the unit an
organization should build and distribute is not a curated agent
configuration bought or bolted on per team, but a hardened harness,
engineered once centrally and run unmodified everywhere.

None of this work addresses the deployment topology argued for here: a
single harness artifact that runs, without modification, as an
unattended container-and-scheduler backbone, as the execution engine
behind a business-facing chat surface, and interactively at a terminal,
under one identity and governance model. That is the gap Section 4
fills.

\section*{3. The Evolution from Chat to
Harness}\label{the-evolution-from-chat-to-harness}
\addcontentsline{toc}{section}{3. The Evolution from Chat to Harness}

Four successive patterns describe how enterprises have integrated
frontier models into their operations since 2022, each addressing a
limitation of the one before it.

\textbf{Chat.} ChatGPT-like SoTA models powered internal chatbots: a
conversation loop ensuring context persists across turns. What it does
not have, figuratively, is ``hands''. It can act only inside the
boundaries of the chat window itself, a file the user uploads, a tool
the user or the app invokes on its behalf, never something it goes and
finds on its own; and the tools available to it are as fixed and narrow
as whatever the product builder anticipated; the model cannot extend
them mid-task. Later versions of various internal chatbots reached
further, out to arbitrary external tools, a pattern and a set of methods
pioneered by LangChain.

\textbf{DAG / chain orchestration.} Graph- and chain-orchestration
frameworks wire multiple model calls into a predefined graph: a designer
decides in advance which node handles which sub-task, and in what order.
This adds persistence and multi-step structure, but at a cost: the model
completes one predefined sub-task per node and can never discover a path
through the problem that the designer did not anticipate. Section 4.2
makes the opposite design choice. The graph is frozen at design time,
and any edge case outside it fails.

\textbf{The autocomplete layer.} A narrower pattern embeds a single
model call directly inside an existing product, as a one-shot, no-hands
suggestion layer: draft this email, summarize this record, suggest this
next field. It is reliable within its very narrow scope, a decoration on
the surrounding product; it proposes text for a human to accept, edit,
or discard.

\textbf{Harness.} A single loop calls the same model repeatedly, feeding
each result back in, so the model can iterate on its own reasoning,
recover from a failed step, and, with a filesystem access, create and
retrieve artifacts of its own work across that loop. The model itself
decides the next step at each iteration, shaping its own working
environment as it goes.

The real work in an enterprise is rarely answering one siloed question.
It is spanning systems, pulling data points from each, and reasoning
over the result iteratively, the same way a human doing the job would.
None of the first three patterns can possibly reach that far.

Nor does driving it require an engineer. A specialist who knows a
problem cold can navigate a model through the one-off system that solves
it: check the direction, nudge, discard, rebuild. A pharmacovigilance
lead who can't code a disproportionality pipeline by hand can still tell
the model to stratify by age when the signal looks wrong; she never
writes a line, but every turn is hers. A terminal is still a barrier for
her. Harnesses are starting to ship a GUI alongside the loop instead, an
Electron app or a thin browser-based interface (microcc includes a thin
TSX interface in the browser).

This paper is not about coding agents. Coding is the domain the harness
mastered first, because models thrive at code generation. The argument
here is that the same paradigm, model in a while loop, one-shot
suggestion replaced by iteration, memory and writing of own methods
plastically addressing a given problem, applies directly to the much
larger body of enterprise activity that is not software engineering at
all: document review, case triage, report generation, and the other
reasoning-and-language tasks that make up the rest of knowledge work.
Section 4 describes the architecture that makes that generalization
operable at enterprise scale.

\section*{4. System Architecture}\label{system-architecture}
\addcontentsline{toc}{section}{4. System Architecture}

Figure 1 previews this section. One harness serves as the backbone,
packaged once. Three deployment surfaces reach it, and four supporting
services back it up; Sections 4.5 to 4.9 explain each in full.

\begin{figure}
\centering
\pandocbounded{\includegraphics[keepaspectratio]{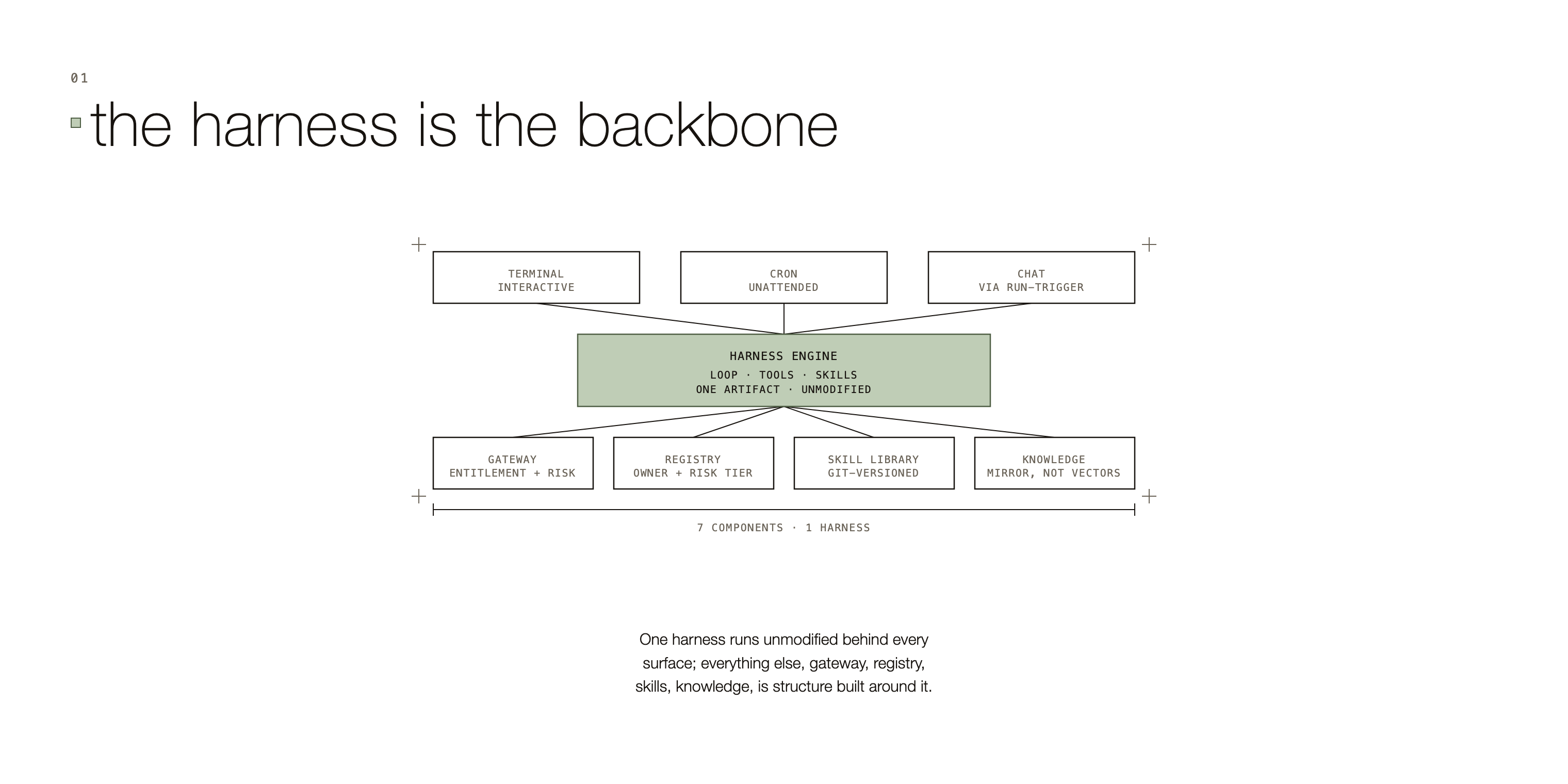}}
\caption{One harness runs unmodified behind every surface; everything
else, gateway, registry, skills, knowledge, is structure built around
it.}
\end{figure}

This section keeps code to a minimum, limited to short illustrative
snippets. We show code only where the underlying mechanism would
otherwise stay abstract: a tool's calling convention, the gateway's
discovery and call path, a solution's configuration file and deployment
template, the registry's registration path, and the run-trigger
endpoint. Numerous open-source harness implementations already exist and
are worth studying directly. Our own work runs on our harness, microcc
(\url{https://pypi.org/project/micro-cc/}).

\subsection*{4.1 Starting condition and design
goals}\label{starting-condition-and-design-goals}
\addcontentsline{toc}{subsection}{4.1 Starting condition and design
goals}

In the large enterprises we studied, numerous teams typically develop
their own custom solutions to these problems: retrieval-augmented
generation systems, code review assistants, requirements-analysis tools,
report generators, most commonly on graph- or chain-based orchestration
frameworks (LangChain-style tooling and its equivalents) or custom
orchestration code. These efforts share no common registry, no common
access layer, and give management no visibility into what exists, who
owns it, or what it costs. Good code is cheap enough to generate now
which leads disparate teams reinventing the wheel, building a custom
solution for the same kind of problem. Using, reviewing, and
understanding these solutions afterward is demanding, because each one
means parsing a different code repository from scratch.

The same architecture has proven capable of running as a cron job,
automating a wide range of back-office work well beyond software
engineering. One example is applying policy rules to a structured record
inside a line-of-business system (e.g.~Dynamics 365 Business Central):
invoking external services as needed and writing the resulting decision
back, so a human review step is automated end to end. Another is
triaging inbound requests against a CRM: pulling context from multiple
connected systems and completing the response, whether that means
drafting correspondence, updating a customer record, or initiating a
downstream transaction. A third, at a manufacturing client, replaces a
manual document check: engineers used to compare a supplier's
certificate of conformance against the required material norm by hand;
the harness now runs that same comparison whatever format the
certificate arrives in, PDF or Excel, against both the SAP export and
the norm itself. In all three cases, extending the automation is a
change to a text file, or an addition to a git-versioned library of such
files.

These examples share a common shape: repeatable but complex tasks that
require reasoning across a variety of systems, exactly where a harness
performs surprisingly well. Claude Code became notable because of a
small set of primitives, the loop and the filesystem introduced in
Section 3, packaged as an architecture that gets more capable on its own
as the underlying models improve, with no redesign required on the
enterprise's part.

Two goals follow from this starting condition. First, enable these teams
to build using models ``with hands'': a model that can loop to iterate
on its own reasoning, recover from mistakes, and explore its own latent
space across steps, and that can create and store artifacts of its work
through filesystem access. Second, give every team the same redeployable
architecture, so no matter what problem a team builds for, the
underlying code stays identical across every deployment. That's what
makes the whole fleet auditable: reviewing a new solution stops being a
code review and becomes reading one instructions file.

\subsection*{4.2 Core design decisions}\label{core-design-decisions}
\addcontentsline{toc}{subsection}{4.2 Core design decisions}

The proposed architecture is based on four decisions.

\textbf{A frontier model as sole orchestrator.} There is no graph or
chain fixed at design time. A single harness instance calls the same
model repeatedly, each iteration observing the result of the last. It
can also spawn further instances of itself, composing its own
orchestration graph at runtime (Section 4.5). Process changes become
text edits to a system prompt rather than development sprints, and model
upgrades improve every deployed solution with zero code changes.

\textbf{A tool gateway as shared capability layer.} One persistent
service holds every enterprise tool integration, a data platform, a
document store, an ERP system behind a single protocol. Teams select
tools from a registry rather than building their own integrations;
adding a new tool is a configuration change on the gateway, instantly
available to every solution.

\textbf{Governance by construction.} Infrastructure policy blocks
non-compliant resources at creation time. CI/CD auto-registers every
deployment in a lifecycle registry. Risk gating happens per call rather
than per tool: the model states \texttt{risky} explicitly on every call,
judged from the specific parameters it just constructed rather than
looked up from a static tool or group definition, and a call omitting it
is rejected outright. A risky call blocks on the judge described in
Section 4.5 before the backend ever runs, and it escalates to human
approval only when that judge itself can't clear it. No change-control
process gets a chance to add a check afterward, because no deployment
path skips these checks to begin with.

\textbf{Simplicity and ease as the path to enterprise-wide adoption:
fork, configure, push.} Whoever owns the problem clones one reference
repository and works it until an instruction file captures what works,
then selects from the company's list of integrations. In practice this
means editing two plain-text files, an instructions file and a
configuration file, and pushing code. CI/CD provisions infrastructure,
deploys, and registers the solution automatically.

\subsection*{4.3 Four architectural
layers}\label{four-architectural-layers}
\addcontentsline{toc}{subsection}{4.3 Four architectural layers}

\begin{itemize}
\tightlist
\item
  \emph{The knowledge substrate} (Section 4.4): a git-mirrored,
  identity-filtered copy of the enterprise's own document stores,
  exposed to the harness to explore and read through directly.
\item
  \emph{The harness engine} (Section 4.5): a loop, memory, and bash\_,
  one reference repository, forkable, driven by two text files (a
  plain-language instructions file and a small configuration file),
  deployed by CI/CD.
\item
  \emph{The tool gateway}, behind the proxy, is the intelligence layer:
  it checks entitlement before executing, and holds a queue for any call
  the model itself declares risky, requiring human sign-off before the
  backend runs.
\item
  \emph{The cloud governance skeleton}: management-group-level policy,
  role-based access control, and tagging; an invisible structure that
  ensures cost transparency and control.
\end{itemize}

\subsection*{4.4 Enterprise ontology is a plain text library: a git
mirror}\label{enterprise-ontology-is-a-plain-text-library-a-git-mirror}
\addcontentsline{toc}{subsection}{4.4 Enterprise ontology is a plain
text library: a git mirror}

We propose a retrieval that is similarly based on a filesystem the
harness already knows how to read.

A retrieval-augmented generation as broadly adopted, means embedding a
query, comparing it against embedded fragments of a document corpus, and
returning the nearest fragments to the model in place of the document
they came from. That's a limiting mechanism at the level frontier models
have reached: retrieval should be active exploration instead. bash\_
(Section 4.5) already gives the harness the tools a human mind would use
to navigate a shared drive. A vectorstore asks it to reason over
similarity scores instead, shutting off the exploration it's otherwise
capable of and discarding the structure, provenance, and surrounding
document a fragment came from. It's also unauditable in a way that
matters at enterprise scale: nothing records which fragments a given run
could have retrieved, so reconstructing what the agent knew when it made
a decision isn't possible after the fact.

We propose a mechanism that mirrors a document source into a plain-text
tree that the harness reads directly. A scheduled sync job, one per
connected source system, walks that system's own native change key (a
delta query against \texttt{cTag} for a SharePoint drive, a page version
for Confluence, a rowversion for a relational source) rather than
re-crawling the corpus on every run, and writes each changed item to a
path that mirrors its position in the source folder tree, so a
document's location in the mirror is a lossless index of where it lives
in the system of record. (Diagrams, document scans, and other visual
documents are transcribed verbatim to text at ingest time through a
vision model.)

One sync cycle:

\begin{Shaded}
\begin{Highlighting}[]
\KeywordTok{async} \KeywordTok{function} \FunctionTok{syncSource}\NormalTok{(source}\OperatorTok{:}\NormalTok{ SourceAdapter}\OperatorTok{,}\NormalTok{ cursor}\OperatorTok{:} \DataTypeTok{string}\NormalTok{)}\OperatorTok{:} \BuiltInTok{Promise}\OperatorTok{\textless{}}\DataTypeTok{string}\OperatorTok{\textgreater{}}\NormalTok{ \{}
  \KeywordTok{const}\NormalTok{ changes }\OperatorTok{=} \ControlFlowTok{await}\NormalTok{ source}\OperatorTok{.}\FunctionTok{delta}\NormalTok{(cursor)}\OperatorTok{;}
  \ControlFlowTok{for}\NormalTok{ (}\KeywordTok{const}\NormalTok{ item }\KeywordTok{of}\NormalTok{ changes) \{}
    \KeywordTok{const}\NormalTok{ text }\OperatorTok{=} \FunctionTok{isTextual}\NormalTok{(item) }\OperatorTok{?}\NormalTok{ item}\OperatorTok{.}\AttributeTok{body} \OperatorTok{:} \ControlFlowTok{await} \FunctionTok{renderToText}\NormalTok{(item)}\OperatorTok{;}
    \ControlFlowTok{await} \FunctionTok{writeMirrorFile}\NormalTok{(}\FunctionTok{mirrorPath}\NormalTok{(source}\OperatorTok{,}\NormalTok{ item)}\OperatorTok{,}\NormalTok{ text}\OperatorTok{,}\NormalTok{ item}\OperatorTok{.}\AttributeTok{aclGroup}\NormalTok{)}\OperatorTok{;}
\NormalTok{  \}}
  \ControlFlowTok{return} \FunctionTok{commitMirror}\NormalTok{(}\VerbatimStringTok{\textasciigrave{}sync: }\SpecialCharTok{$\{}\NormalTok{source}\OperatorTok{.}\AttributeTok{id}\SpecialCharTok{\}}\VerbatimStringTok{ @ }\SpecialCharTok{$\{}\NormalTok{changes}\OperatorTok{.}\AttributeTok{newCursor}\SpecialCharTok{\}}\VerbatimStringTok{\textasciigrave{}}\NormalTok{)}\OperatorTok{;}
\NormalTok{\}}
\end{Highlighting}
\end{Shaded}

The mirror's own git history is what makes it auditable. Every sync
cycle is one commit. The mirror's state at any past moment is a
specific, addressable commit. Each read the model makes against the
mirror resolves against whichever commit is current at that moment, and
the gateway logs that commit alongside the read itself. An auditor
asking what a solution could have known at the moment of a given
decision pulls that run's logged reads and checks out the commits they
resolved against. That reconstructs the exact content the model had in
front of it at that moment, even if it was several sync cycles before
the mirror's current state.

\texttt{mirrorPath} writes each file's source ACL alongside it as a
directory-group reference. The read tool the model calls resolves
against the caller's own groups before returning content. This is the
same discovery-then-filter path Section 4.6 already applies to the tool
catalog. A caller outside a document's ACL group doesn't see it listed,
let alone read it. One mechanism drives both the knowledge substrate and
the tool catalog.

\subsection*{4.5 The harness engine}\label{the-harness-engine}
\addcontentsline{toc}{subsection}{4.5 The harness engine}

The engine is the loop itself: one process that calls the model
repeatedly, feeds each result back in, and gives it memory and
filesystem access across iterations. Everything in this section runs on
top of that one primitive.

\textbf{A harness can write and spawn harness.} The same primitives, a
loop, memory, a filesystem, bash, are what allows harness to write a
graph plastic to the problem it is solving and spawn multiple harnesses
- more of itself. It runs the same harness again, headless, pointed at a
fresh folder, and reads the outcome back off a file it writes to. A lead
agent spawns and coordinates worker agents, each exploring its own slice
of a problem, and collects their results back
(\url{https://www.anthropic.com/engineering/multi-agent-research-system}).
In our implementation, microcc's version tracks each spawned child
through a small status file instead of polling it, and children message
each other through the same filesystem: a live socket if the target is
running, a durable inbox if it isn't. The graph is not fixed. Given a
problem, the model decides at runtime the hierarchy, structure, sequence
and prompt to spawn instances of itself with.

A single engine provides the features below, all fairly familiar by now.

\emph{The loop.} One system message is built once per run and never
rewritten; only the surrounding context is rebuilt each iteration.

\emph{Instructions (Skills).} Individually loadable documents the model
pulls in only when it decides they're relevant, accompanied by whatever
scripts and other artifacts they need. What matters at enterprise scale
is where those documents load from. Skills resolve across three tiers:
built-in, shared platform library, and project-specific, with later
tiers overriding earlier ones on a name clash. A central team maintains
one versioned procedural library this way, independent of any team's own
fork. The shared library is a real git repository. The loader
authenticates against it using the engineer's own personal access token.
That's the same credential-scoped principle Section 4.5 applies to tool
calls, so library entitlement is just another repository permission to
administer. Stripped to its essentials, the loader's own tier-resolution
path looks like this:

\begin{Shaded}
\begin{Highlighting}[]
\KeywordTok{async} \KeywordTok{function} \FunctionTok{loadSkills}\NormalTok{(project}\OperatorTok{:}\NormalTok{ ProjectContext}\OperatorTok{,}\NormalTok{ pat}\OperatorTok{:} \DataTypeTok{string}\NormalTok{)}\OperatorTok{:} \BuiltInTok{Promise}\OperatorTok{\textless{}}\NormalTok{SkillIndex}\OperatorTok{\textgreater{}}\NormalTok{ \{}
  \KeywordTok{const}\NormalTok{ tiers }\OperatorTok{=} \ControlFlowTok{await} \BuiltInTok{Promise}\OperatorTok{.}\FunctionTok{all}\NormalTok{([}
    \FunctionTok{loadFromDir}\NormalTok{(BUILTIN\_SKILLS\_DIR)}\OperatorTok{,}                        \CommentTok{// shipped with the harness}
    \FunctionTok{loadFromGit}\NormalTok{(SKILL\_LIBRARY\_REPO}\OperatorTok{,}\NormalTok{ pat}\OperatorTok{,}\NormalTok{ project}\OperatorTok{.}\AttributeTok{selected}\NormalTok{)}\OperatorTok{,} \CommentTok{// shared platform library}
    \FunctionTok{loadFromDir}\NormalTok{(path}\OperatorTok{.}\FunctionTok{join}\NormalTok{(project}\OperatorTok{.}\AttributeTok{root}\OperatorTok{,} \StringTok{"skills"}\NormalTok{))}\OperatorTok{,}         \CommentTok{// project{-}specific, wins on clash}
\NormalTok{  ])}\OperatorTok{;}
  \ControlFlowTok{return}\NormalTok{ tiers}\OperatorTok{.}\FunctionTok{reduce}\NormalTok{((index}\OperatorTok{,}\NormalTok{ tier) }\KeywordTok{=\textgreater{}}\NormalTok{ (\{ }\OperatorTok{...}\NormalTok{index}\OperatorTok{,} \OperatorTok{...}\NormalTok{tier \})}\OperatorTok{,}\NormalTok{ \{\})}\OperatorTok{;}
\NormalTok{\}}
\end{Highlighting}
\end{Shaded}

\texttt{loadFromGit} resolves against the engineer's own PAT while
working interactively. That's the same identity that gates every other
tool call in Section 4.6. At deploy time, this same selection becomes a
build-time decision by default. A solution's \texttt{config.yaml} names
which skills from the library it wants under \texttt{skills:} (Section
4.7), and CI/CD fetches exactly those into the image it builds. The
running container holds no library credential of its own, so a deployed
solution's skill set is baked into the image it was built from and fixed
there. A compromised runtime identity can't expand it by reading further
into the library.

An engineer can also name a skill under \texttt{live\_skills:} in the
config.yaml file (Section 4.7). For a live skill, CI/CD provisions that
deployment's own scoped, identity-bound token to the skill library
repository. This is the same on-behalf-of exchange Section 4.5 already
uses to resolve a tool's credential. The running container then calls
\texttt{loadFromGit} itself at the start of each run, a direct HTTPS
request to GitHub rather than a read from the image, and picks up
whatever commit the named skill is at in that moment.

\begin{Shaded}
\begin{Highlighting}[]
\KeywordTok{async} \KeywordTok{function} \FunctionTok{resolveSkill}\NormalTok{(name}\OperatorTok{:} \DataTypeTok{string}\OperatorTok{,}\NormalTok{ mode}\OperatorTok{:} \StringTok{"baked"} \OperatorTok{|} \StringTok{"live"}\OperatorTok{,}\NormalTok{ token}\OperatorTok{?:}\NormalTok{ ScopedToken)}\OperatorTok{:} \BuiltInTok{Promise}\OperatorTok{\textless{}}\NormalTok{Skill}\OperatorTok{\textgreater{}}\NormalTok{ \{}
  \ControlFlowTok{if}\NormalTok{ (mode }\OperatorTok{===} \StringTok{"baked"}\NormalTok{) }\ControlFlowTok{return} \FunctionTok{readFromImage}\NormalTok{(name)}\OperatorTok{;}       \CommentTok{// read once, at run start, no network call}
  \ControlFlowTok{return} \FunctionTok{fetchFromGit}\NormalTok{(SKILL\_LIBRARY\_REPO}\OperatorTok{,}\NormalTok{ name}\OperatorTok{,}\NormalTok{ token}\OperatorTok{!}\NormalTok{)}\OperatorTok{;}  \CommentTok{// direct HTTPS request to GitHub, every run}
\NormalTok{\}}
\end{Highlighting}
\end{Shaded}

\emph{Tool discovery.} A small always-on set of tools (file operations,
memory, skill-reading) ships with every run. Everything else is
discovered from the tool gateway at runtime, resolved once per run and
cached. That discovery call resolves against the engineer's own account
interactively, the deployment's own identity on a cron schedule
(provisioned at deploy time and granted the directory groups its
configuration named, Section 4.7), or the end user's own identity behind
a chat-triggered run (carried through the run-trigger service, Section
4.9). The catalog surfaced to any of these is already filtered
server-side by the gateway's entitlement rules (Section 4.6).

\emph{bash\_ is all you need.} Frontier models are trained heavily on
shell usage. Reading a man page or an API's own documentation and
composing the right invocation from it is a capability native to the
model. bash\_ is where that training shows up most visibly, general
enough that no enterprise integration work is required for a model to
use it well. We therefore propose a mechanism that bets on this fluency,
composing a correct call from documentation and general knowledge rather
than from a strict method built in advance.

\emph{Credential-scoped tooling: tools as token exchanges.} This is the
mechanism named in Section 1. bash\_ is the one tool that's always
present, and every other tool in the catalog is trying to be more like
it. A curated tool catalog, one pre-built method per anticipated
operation, doesn't scale to the number of enterprise systems a team will
need, and it freezes the model's access to whatever a developer happened
to anticipate at integration time. For a system like SharePoint, the
gateway doesn't register a set of methods like list-files or get-item.
It registers a \textbf{credential broker} instead: it authenticates the
caller's identity against the target system and hands the harness a
single, narrowly scoped token. Against that token, the model then calls
one generic request tool, the same shape as bash\_ itself. It composes
its own Graph query, OData filter, or search syntax, drawing on the same
general knowledge of these APIs it would draw on to compose a shell
command. Identity and entitlement (Section 4.6) decide which credential,
and so which system, the model gets access to. Once it has that
credential, nothing limits which specific calls it can make within that
system.

This design bets on what frontier models already carry from training: a
strong prior on the shape of major enterprise APIs (Microsoft Graph, SAP
OData, Databricks REST and SQL, Sparx EA's schema), built up from the
sheer volume of public documentation and code covering them. from
experience, we argue that a strict, per-action JSON Schema with typed,
enumerated parameters (this includes broadly MCP) is a net negative. It
substitutes a narrower, hand-maintained model of the API for a broader
one the model already has, and freezes access to whatever operations the
schema's author happened to anticipate. In the harness paradigm the
model explores a system it's handed the same way it would explore one
from a terminal, composing a request from documentation and general
knowledge instead of selecting from an enumerated list. The tool surface
that fits it best is the one that gets out of the way: a single generic
request tool per backend, with \texttt{params} left open. Against a SQL
warehouse, this same tool takes \texttt{\{sql,\ catalog,\ schema?\}}
instead of Graph's \texttt{\{method,\ path,\ body?\}}. That's a
different shape for a different backend, and the model picks it from
what it already knows about that backend.

The model-facing envelope around that tool is
\texttt{\{params:\ object,\ risky:\ boolean\}}. The model states
\texttt{risky} explicitly on every call, judged from the specific
\texttt{params} it just constructed as a required parameter; a call that
omits it is rejected before it reaches a backend (Section 4.6), with an
error message that forces the model to ``re-think.'' Risk, in other
words, is call-scoped and self-declared, never a static tier attached
top-down to a tool or a tool group.

As a rough draft, the calling method for such a tool can look like this:

\begin{Shaded}
\begin{Highlighting}[]
\KeywordTok{class}\NormalTok{ GraphTool }\KeywordTok{implements}\NormalTok{ Tool \{}
  \KeywordTok{private}\NormalTok{ token}\OperatorTok{:}\NormalTok{ CachedToken }\OperatorTok{|} \DataTypeTok{null} \OperatorTok{=} \KeywordTok{null}\OperatorTok{;}

  \KeywordTok{async} \FunctionTok{execute}\NormalTok{(call}\OperatorTok{:}\NormalTok{ \{ method}\OperatorTok{:}\NormalTok{ HttpMethod}\OperatorTok{;}\NormalTok{ path}\OperatorTok{:} \DataTypeTok{string}\OperatorTok{;}\NormalTok{ body}\OperatorTok{?:} \DataTypeTok{unknown}\NormalTok{ \}) \{}
    \KeywordTok{const}\NormalTok{ token }\OperatorTok{=} \ControlFlowTok{await} \KeywordTok{this}\OperatorTok{.}\FunctionTok{resolveToken}\NormalTok{()}\OperatorTok{;}
    \ControlFlowTok{return} \KeywordTok{this}\OperatorTok{.}\AttributeTok{client}\OperatorTok{.}\FunctionTok{request}\NormalTok{(\{}
\NormalTok{      method}\OperatorTok{:}\NormalTok{ call}\OperatorTok{.}\AttributeTok{method}\OperatorTok{,}
\NormalTok{      url}\OperatorTok{:} \VerbatimStringTok{\textasciigrave{}}\SpecialCharTok{$\{}\NormalTok{GRAPH\_BASE}\SpecialCharTok{\}$\{}\NormalTok{call}\OperatorTok{.}\AttributeTok{path}\SpecialCharTok{\}}\VerbatimStringTok{\textasciigrave{}}\OperatorTok{,}
\NormalTok{      headers}\OperatorTok{:}\NormalTok{ \{ Authorization}\OperatorTok{:} \VerbatimStringTok{\textasciigrave{}Bearer }\SpecialCharTok{$\{}\NormalTok{token}\OperatorTok{.}\AttributeTok{value}\SpecialCharTok{\}}\VerbatimStringTok{\textasciigrave{}}\NormalTok{ \}}\OperatorTok{,}
\NormalTok{      data}\OperatorTok{:}\NormalTok{ call}\OperatorTok{.}\AttributeTok{body}\OperatorTok{,}
\NormalTok{    \})}\OperatorTok{;}
\NormalTok{  \}}

  \KeywordTok{private} \KeywordTok{async} \FunctionTok{resolveToken}\NormalTok{()}\OperatorTok{:} \BuiltInTok{Promise}\OperatorTok{\textless{}}\NormalTok{CachedToken}\OperatorTok{\textgreater{}}\NormalTok{ \{}
    \ControlFlowTok{if}\NormalTok{ (}\KeywordTok{this}\OperatorTok{.}\AttributeTok{token} \OperatorTok{\&\&} \OperatorTok{!}\FunctionTok{isExpired}\NormalTok{(}\KeywordTok{this}\OperatorTok{.}\AttributeTok{token}\NormalTok{)) }\ControlFlowTok{return} \KeywordTok{this}\OperatorTok{.}\AttributeTok{token}\OperatorTok{;}
    \KeywordTok{this}\OperatorTok{.}\AttributeTok{token} \OperatorTok{=} \ControlFlowTok{await} \FunctionTok{exchangeOnBehalfOf}\NormalTok{(}\KeywordTok{this}\OperatorTok{.}\AttributeTok{callerIdentity}\OperatorTok{,}\NormalTok{ GRAPH\_SCOPE)}\OperatorTok{;}
    \ControlFlowTok{return} \KeywordTok{this}\OperatorTok{.}\AttributeTok{token}\OperatorTok{;}
\NormalTok{  \}}
\NormalTok{\}}
\end{Highlighting}
\end{Shaded}

\texttt{execute} takes a method, a path, and a body, the same three
free-form parameters bash\_ takes as a single command string. There's no
fixed enum of supported operations. \texttt{resolveToken} runs first on
every call, refreshing an expired token before the request executes. The
model never sees the token or the exchange, only the result of the call
it composed. There's no per-operation method on this class to audit,
because there isn't one to add.

Each backend still contributes a short, hand-written
\texttt{usage\_hint} string, merged with \texttt{tools.yaml}'s own
\texttt{description} at discovery time and surfaced to the model as that
tool's docstring:

\begin{Shaded}
\begin{Highlighting}[]
\KeywordTok{async} \KeywordTok{function} \FunctionTok{describeTool}\NormalTok{(id}\OperatorTok{:} \DataTypeTok{string}\NormalTok{)}\OperatorTok{:} \BuiltInTok{Promise}\OperatorTok{\textless{}}\NormalTok{ToolDescriptor}\OperatorTok{\textgreater{}}\NormalTok{ \{}
  \KeywordTok{const}\NormalTok{ backend }\OperatorTok{=}\NormalTok{ registry}\OperatorTok{.}\FunctionTok{get}\NormalTok{(id)}\OperatorTok{;}
  \KeywordTok{const}\NormalTok{ declared }\OperatorTok{=}\NormalTok{ toolsYaml}\OperatorTok{.}\FunctionTok{get}\NormalTok{(id)}\OperatorTok{;} \CommentTok{// description, directoryGroups}
  \ControlFlowTok{return}\NormalTok{ \{}
    \OperatorTok{...}\NormalTok{declared}\OperatorTok{,}
\NormalTok{    description}\OperatorTok{:} \VerbatimStringTok{\textasciigrave{}}\SpecialCharTok{$\{}\NormalTok{declared}\OperatorTok{.}\AttributeTok{description}\SpecialCharTok{\}\textbackslash{}n\textbackslash{}n$\{}\NormalTok{backend}\OperatorTok{.}\FunctionTok{usageHint}\NormalTok{()}\SpecialCharTok{\}}\VerbatimStringTok{\textasciigrave{}}\OperatorTok{,}
\NormalTok{  \}}\OperatorTok{;}
\NormalTok{\}}
\end{Highlighting}
\end{Shaded}

\texttt{usageHint()} doesn't re-teach the backend's public API, since
the model already knows the common conventions of Graph, OData, SQL, and
the like cold. Instead it supplies this particular deployment's own
private wiring. For a backend exposing both a SQL and a REST surface,
that means stating which of the two valid \texttt{params} shapes
applies, \texttt{\{sql,\ catalog,\ schema?\}} or
\texttt{\{method,\ path,\ body?\}}. When a backend has more than one
runtime mode, the hint names the one currently live, an operational fact
the model has no way to infer on its own. The backend, in other words,
tells the model how this particular deployment packages a request it
already knows how to write. Training supplies the model's knowledge of
the API itself. The hint supplies just enough local context to aim that
knowledge correctly. The credential, supplied by the backend, is never
exposed to the model. There's no static schema validation catching a
malformed call before it reaches the backend under this design. A wrong
\texttt{params} shape just fails loudly there. The gateway does validate
the self-declared \texttt{risky} flag, which is required on every call.

\emph{Three kinds of state} Conversation history is durable, so a run
can pick back up after a crash or an incomplete session. A short-lived,
TTL-bound cache holds the tool-discovery result, shared across every
replica of a running solution. Long-term memory is scoped to the
caller's identity, so it persists across runs and even across different
solutions. The instructions file puts that last kind of state to work.
It tells the model to write down what it learns as it goes, such as a
workaround that held or a constraint it needs to remember, as a keyed
memory entry (another simple Anthropic primitive), so the next run
doesn't have to rediscover it. Only the keys get loaded into context on
each loop iteration, not the full entries. That's the same lazy-load
pattern Skills already use above. A lesson from a past run costs nothing
until the model actually decides it's worth reading in full.

\emph{The judge: a spawned verdict, not a self-report.} A call the model
declares \texttt{risky} needs review before it reaches a human. Left to
the model that made the call, the same context that composed it is also
the one grading it, exactly the setup that lets a constraint drop
somewhere in a long reasoning trace (arXiv:2604.13107). We route it
through the sub-harness mechanism described above instead. The flagged
call spawns a fresh sub-harness, asked to review the call, the calling
solution's instructions file, and its trace so far. All along, the
harnesses communicate the same way any two spawned instances do: no API
between them, just a shared filesystem and an \texttt{inbox.json} each
one writes to and polls.

\emph{Approval gate for tool calls model self-declares as risky.} A call
the model has declared \texttt{risky} first goes to the judge just
described. Only a verdict the judge itself can't resolve, or clears as
still risky, reaches a human. \texttt{approvalQueue.enqueueAndWait}
(Section 4.6) is what actually blocks at that point. It posts the call,
its \texttt{params}, the judge's own reasoning, and the caller's
identity to, for example, a Teams channel via webhook, with one plain
approve/deny action per message. It holds the request open rather than
returning until a named approver on that channel clicks one. The click
hits the gateway's own decision endpoint directly. This is the same
pattern Section 4.8 uses for registration review, and it resolves the
pending call and lets \texttt{enqueueAndWait} return.

\emph{Self-reporting.} Every run ends with the model stating its own
outcome in one structured status line: done, needs input, or failed,
plus a summary. The surrounding wrapper extracts this line itself rather
than relying on a separate classifier call, and posts it to the
lifecycle registry. The registry persists it alongside the solution's
owner, the problem it addresses, and its run history.

\subsection*{4.6 Identity and tool
visibility}\label{identity-and-tool-visibility}
\addcontentsline{toc}{subsection}{4.6 Identity and tool visibility}

Identity decides tool visibility. A solution's config.yaml declares
which tool groups it wants. CI/CD reads that once, to request the
solution's deployed identity be added to the matching directory group.
At runtime, the gateway's entitlement mapping, plain YAML, admin-edited,
git-versioned, is the only thing deciding what a caller can see and
call. The caller could be an unattended deployment's own identity, an
interactive engineer's account, or an end user's identity behind a
chat-triggered run (Section 4.9); the gateway checks all three the same
way. A caller sees only the tools its groups map to at discovery time.
Anything outside that 404s on describe and 403s on call. The gateway's
discovery and call path:

\begin{Shaded}
\begin{Highlighting}[]
\KeywordTok{async} \KeywordTok{function} \FunctionTok{discover}\NormalTok{(caller}\OperatorTok{:}\NormalTok{ Identity)}\OperatorTok{:} \BuiltInTok{Promise}\OperatorTok{\textless{}}\NormalTok{ToolCatalog}\OperatorTok{\textgreater{}}\NormalTok{ \{}
  \KeywordTok{const}\NormalTok{ groups }\OperatorTok{=} \ControlFlowTok{await} \FunctionTok{resolveDirectoryGroups}\NormalTok{(caller)}\OperatorTok{;}
  \ControlFlowTok{return}\NormalTok{ catalog}\OperatorTok{.}\FunctionTok{filter}\NormalTok{(tool }\KeywordTok{=\textgreater{}}\NormalTok{ groups}\OperatorTok{.}\FunctionTok{some}\NormalTok{(g }\KeywordTok{=\textgreater{}}\NormalTok{ tool}\OperatorTok{.}\AttributeTok{directoryGroups}\OperatorTok{.}\FunctionTok{has}\NormalTok{(g)))}\OperatorTok{;}
\NormalTok{\}}

\KeywordTok{async} \KeywordTok{function} \FunctionTok{call}\NormalTok{(caller}\OperatorTok{:}\NormalTok{ Identity}\OperatorTok{,}\NormalTok{ toolId}\OperatorTok{:} \DataTypeTok{string}\OperatorTok{,}\NormalTok{ request}\OperatorTok{:}\NormalTok{ ToolRequest) \{}
  \KeywordTok{const}\NormalTok{ tool }\OperatorTok{=}\NormalTok{ (}\ControlFlowTok{await} \FunctionTok{discover}\NormalTok{(caller))}\OperatorTok{.}\FunctionTok{get}\NormalTok{(toolId)}\OperatorTok{;}
  \ControlFlowTok{if}\NormalTok{ (}\OperatorTok{!}\NormalTok{tool) }\ControlFlowTok{return}\NormalTok{ \{ status}\OperatorTok{:} \DecValTok{403}\NormalTok{ \}}\OperatorTok{;} \CommentTok{// exists in the catalog, just not for this caller}
  \ControlFlowTok{if}\NormalTok{ (}\KeywordTok{typeof}\NormalTok{ request}\OperatorTok{.}\AttributeTok{risky} \OperatorTok{!==} \StringTok{"boolean"}\NormalTok{) }\ControlFlowTok{return}\NormalTok{ \{ status}\OperatorTok{:} \DecValTok{400}\NormalTok{ \}}\OperatorTok{;} \CommentTok{// no default, no inference}
  \ControlFlowTok{if}\NormalTok{ (request}\OperatorTok{.}\AttributeTok{risky}\NormalTok{) \{}
    \KeywordTok{const}\NormalTok{ review }\OperatorTok{=} \ControlFlowTok{await}\NormalTok{ judge}\OperatorTok{.}\FunctionTok{spawnAndWait}\NormalTok{(caller}\OperatorTok{,}\NormalTok{ toolId}\OperatorTok{,}\NormalTok{ request)}\OperatorTok{;} \CommentTok{// fresh sub{-}harness, one verdict}
    \ControlFlowTok{if}\NormalTok{ (review}\OperatorTok{.}\AttributeTok{status} \OperatorTok{!==} \StringTok{"resolved"}\NormalTok{) \{}
      \KeywordTok{const}\NormalTok{ verdict }\OperatorTok{=} \ControlFlowTok{await}\NormalTok{ approvalQueue}\OperatorTok{.}\FunctionTok{enqueueAndWait}\NormalTok{(caller}\OperatorTok{,}\NormalTok{ toolId}\OperatorTok{,}\NormalTok{ request}\OperatorTok{,}\NormalTok{ review}\OperatorTok{.}\AttributeTok{reasoning}\NormalTok{)}\OperatorTok{;}
      \ControlFlowTok{if}\NormalTok{ (verdict }\OperatorTok{!==} \StringTok{"approved"}\NormalTok{) }\ControlFlowTok{return}\NormalTok{ \{ status}\OperatorTok{:} \DecValTok{403}\NormalTok{ \}}\OperatorTok{;}
\NormalTok{    \}}
\NormalTok{  \}}
  \ControlFlowTok{return}\NormalTok{ tool}\OperatorTok{.}\FunctionTok{execute}\NormalTok{(request}\OperatorTok{.}\AttributeTok{params}\NormalTok{)}\OperatorTok{;}
\NormalTok{\}}
\end{Highlighting}
\end{Shaded}

\texttt{discover} is what the harness calls once per run to build the
catalog the model sees. It also serves as the describe path. A tool
absent from a caller's filtered result 404s. \texttt{call} re-derives
that same filtered set instead of trusting whatever tool id the model
sends, so a tool id outside it 403s. That's the caller's own discovery
result applied a second time. There's no separate permissions check to
keep in sync. Risk gating in \texttt{call} isn't a second lookup against
that same filtered set.

\texttt{risky} is a required field on the request. The model sets it
from its own judgment of the specific \texttt{params} it just
constructed. A call that omits it 400s before either the entitlement
check or the judge (Section 4.5) runs. The harness doesn't filter tools,
hide them, or assign risk levels itself.

The same harness artifact is safe to run unmodified in any of the three
contexts described in Section 4.10. An unattended deployment's own
identity, an interactive engineer's own identity, and a chat-triggered
run's end-user identity all pass through exactly the same check. There's
no second, harness-side copy of this logic to keep in sync or to bypass.

\subsection*{4.7 Deployment mechanism: fork, configure,
push}\label{deployment-mechanism-fork-configure-push}
\addcontentsline{toc}{subsection}{4.7 Deployment mechanism: fork,
configure, push}

Deployment spans five steps. An engineer first runs the harness
interactively, live from a terminal, against the real problem, with no
configuration written yet. Tool discovery already shows only what their
own identity is entitled to see, so there's nothing to request before
this first pass. The engineer then distills what worked into a
plain-language instructions file. Anything reusable beyond the one case
is split out into a separate skill document, kept out of the
instructions themselves. A handful of fields (owner, a one-line problem
statement, target tool groups, business area) go into the configuration
file, alongside ordinary settings such as model choice, schedule, and
trigger mode:

\begin{Shaded}
\begin{Highlighting}[]
\FunctionTok{owner}\KeywordTok{:}\AttributeTok{ crm{-}team@company.com}
\FunctionTok{problem}\KeywordTok{:}\AttributeTok{ }\StringTok{"Triage inbound CRM requests and draft first{-}pass responses"}
\FunctionTok{tool\_groups}\KeywordTok{:}\AttributeTok{ }\KeywordTok{[}\AttributeTok{crm{-}triage}\KeywordTok{,}\AttributeTok{ sharepoint{-}readonly}\KeywordTok{]}
\FunctionTok{skills}\KeywordTok{:}\AttributeTok{ }\KeywordTok{[}\AttributeTok{crm{-}triage{-}playbook}\KeywordTok{,}\AttributeTok{ sharepoint{-}search{-}patterns}\KeywordTok{]}
\FunctionTok{live\_skills}\KeywordTok{:}\AttributeTok{ }\KeywordTok{[}\AttributeTok{crm{-}triage{-}playbook}\KeywordTok{]}\CommentTok{   \# fetched from GitHub at each run instead of baked in}
\FunctionTok{business\_area}\KeywordTok{:}\AttributeTok{ customer{-}service}
\FunctionTok{model}\KeywordTok{:}\AttributeTok{ claude{-}sonnet{-}5}
\FunctionTok{schedule}\KeywordTok{:}\AttributeTok{ }\StringTok{"*/15 * * * *"}
\FunctionTok{trigger}\KeywordTok{:}\AttributeTok{ cron}
\FunctionTok{registration\_id}\KeywordTok{:}\CommentTok{   \# {-}\textgreater{} left blank; filled in by CI/CD on first deploy}
\end{Highlighting}
\end{Shaded}

On push, CI/CD reads the configuration. If it carries no registration
id, this is a first deploy: CI/CD registers the solution with the
lifecycle registry (a GitHub Action step that runs the overlap and risk
checks described in Section 4.8 before anything is provisioned),
surfaces anything worth a human's attention in the same build summary
the engineer is already reading, and commits the returned id back into
the configuration file as a bot commit. Every subsequent push instead
updates the existing registration. Building the image is also where the
configuration's \texttt{skills} field (Section 4.5) is resolved. A
GitHub Action step pulls those documents from the skill library, using a
service credential scoped to that repository rather than the engineer's
own PAT, and bakes them into the image alongside the harness engine
itself. Any name also listed under \texttt{live\_skills} is handled
differently. CI/CD instead provisions that deployment's own narrowly
scoped, identity-bound token to the library repository, so the running
container reaches GitHub directly at each run rather than relying on
what was baked in (Section 4.5). A deployed container therefore carries
no library credential beyond what its own configuration explicitly named
live, and can't reach the library at runtime for anything beyond that.
CI/CD then builds and pushes the container image(s) and generates
infrastructure-as-code from a shared module plus the configuration's own
fields:

\begin{Shaded}
\begin{Highlighting}[]
\NormalTok{module "solution" \{}
\NormalTok{  source      = "../modules/harness{-}solution"}
\NormalTok{  owner       = var.owner}
\NormalTok{  tool\_groups = var.tool\_groups}
\NormalTok{  schedule    = var.schedule}
\NormalTok{  trigger     = var.trigger}
\NormalTok{\}}
\end{Highlighting}
\end{Shaded}

Every solution's deployment template is this one module call with a
different set of arguments. The module itself encodes the
governance-by-construction decisions from Section 4.2, not the
configuration file. CI/CD applies it, then requests the directory group
membership that grants runtime tool visibility. It can only do this once
real infrastructure exists, since a deployed identity's directory object
doesn't exist before that. The solution then runs unattended, on its
schedule or trigger, through the same engine. It's gated by the same
entitlement and approval mechanisms as the interactive path.

The engineer never touches authentication, model client code,
infrastructure scripts, audit logging, tool implementations, tool
filtering, or the registration call itself; all of it is structural,
fixed once for every solution.

Figure 3 walks one specialist through these five steps. An
accounts-payable analyst forks the reference repository and runs it live
against three-way-match exceptions in Dynamics 365 Business Central:
purchase order, receipt, and invoice lines that fail to reconcile within
policy tolerance. Watching the model compose the OData queries and
tolerance logic that resolve a batch of these by hand, she distills the
pattern into a skill document once it holds up across enough cases. She
commits it to the shared skill library under her own PAT. This is the
same credential-scoped path Section 4.5 describes. Naming that skill and
a cron schedule in the configuration file, then pushing, is the only
remaining step that's hers. CI/CD bakes the named skill into the image
by default, or, had she named it under \texttt{live\_skills} instead,
provisions the deployment's own scoped token so the container fetches it
from GitHub at each run. It also registers the solution and deploys it.
From that point, the same unmodified harness re-reads her skill
unattended, on schedule, against every new exception batch.

\begin{figure}
\centering
\pandocbounded{\includegraphics[keepaspectratio]{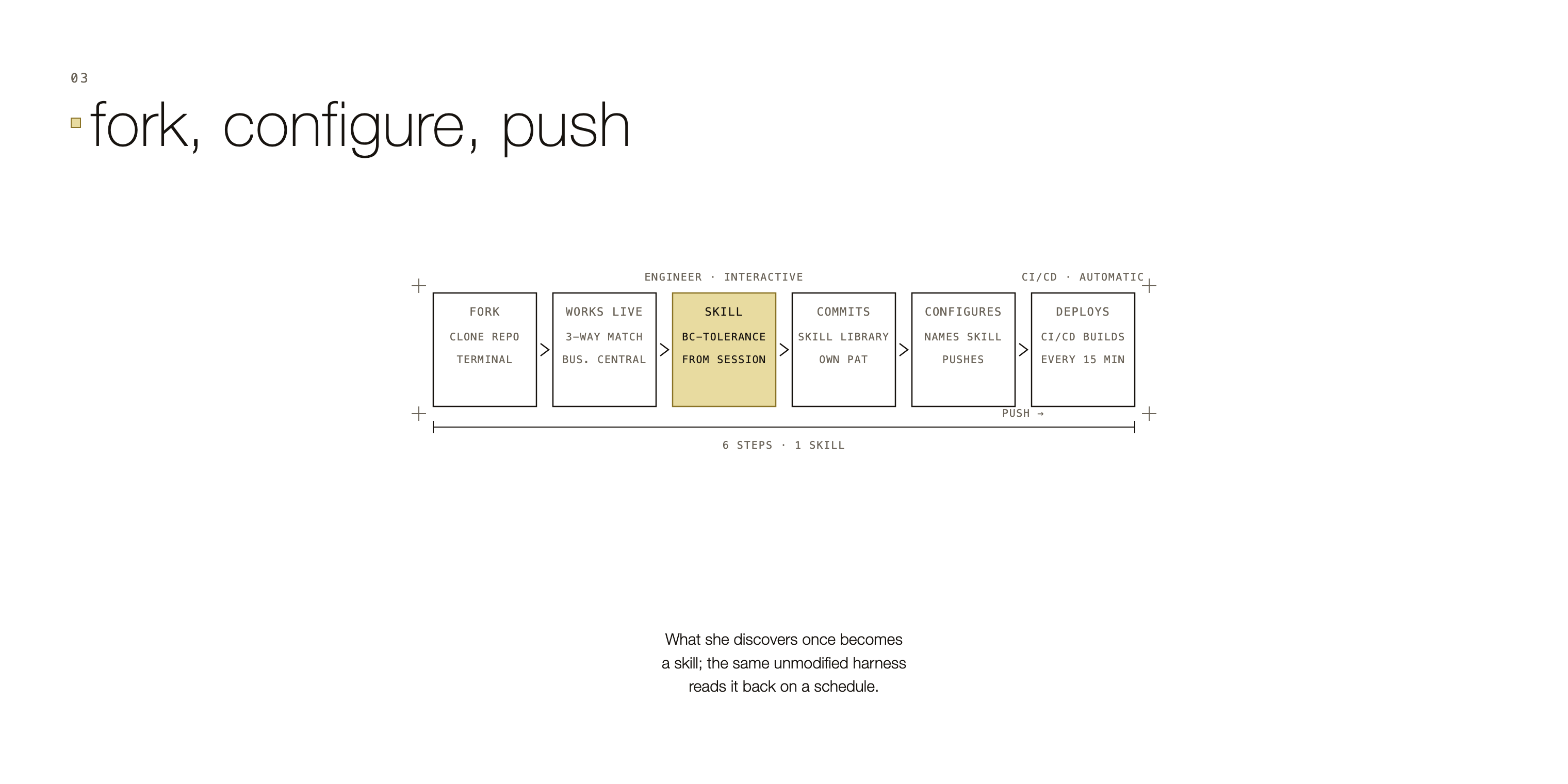}}
\caption{A specialist's fork moves through six steps, from an
interactive terminal session to an unattended cron deployment; the skill
she writes is the one artifact that survives the push.}
\end{figure}

\subsection*{4.8 Registration gating}\label{registration-gating}
\addcontentsline{toc}{subsection}{4.8 Registration gating}

In our proposed mechanism, registration happens naturally, as a side
effect of pushing code (Section 4.7). CI/CD resolves the directory group
memberships of the human who triggered the deploy through one directory
API call, using their already-real single-sign-on identity in place of a
service account. A separate entitlement gate, a small git-managed file
using the same idiom as the tool gateway's entitlement mapping, checks
whether this person is allowed to register a solution in this business
area at all. Simplified snippet:

\begin{Shaded}
\begin{Highlighting}[]
\KeywordTok{async} \KeywordTok{function} \FunctionTok{register}\NormalTok{(deploy}\OperatorTok{:}\NormalTok{ DeployEvent)}\OperatorTok{:} \BuiltInTok{Promise}\OperatorTok{\textless{}}\NormalTok{RegistrationResult}\OperatorTok{\textgreater{}}\NormalTok{ \{}
  \KeywordTok{const}\NormalTok{ requester }\OperatorTok{=} \ControlFlowTok{await} \FunctionTok{resolveDirectoryGroups}\NormalTok{(deploy}\OperatorTok{.}\AttributeTok{triggeredBy}\NormalTok{)}\OperatorTok{;}
  \ControlFlowTok{if}\NormalTok{ (}\OperatorTok{!}\NormalTok{entitlementGate}\OperatorTok{.}\FunctionTok{allows}\NormalTok{(requester}\OperatorTok{,}\NormalTok{ deploy}\OperatorTok{.}\AttributeTok{businessArea}\NormalTok{)) \{}
    \ControlFlowTok{return} \FunctionTok{pendingReview}\NormalTok{(deploy}\OperatorTok{,} \StringTok{"entitlement"}\NormalTok{)}\OperatorTok{;}
\NormalTok{  \}}
  \KeywordTok{const}\NormalTok{ overlap }\OperatorTok{=} \ControlFlowTok{await} \FunctionTok{findSimilar}\NormalTok{(deploy)}\OperatorTok{;}
  \ControlFlowTok{if}\NormalTok{ (overlap}\OperatorTok{.}\AttributeTok{score} \OperatorTok{\textgreater{}}\NormalTok{ BLOCK\_THRESHOLD) }\ControlFlowTok{return} \FunctionTok{pendingReview}\NormalTok{(deploy}\OperatorTok{,} \StringTok{"overlap"}\NormalTok{)}\OperatorTok{;}
  \KeywordTok{const}\NormalTok{ risk }\OperatorTok{=} \ControlFlowTok{await} \FunctionTok{assessRisk}\NormalTok{(deploy)}\OperatorTok{;}
  \ControlFlowTok{if}\NormalTok{ (risk}\OperatorTok{.}\AttributeTok{risky}\NormalTok{) }\ControlFlowTok{return} \FunctionTok{pendingReview}\NormalTok{(deploy}\OperatorTok{,} \StringTok{"risk"}\NormalTok{)}\OperatorTok{;}
  \ControlFlowTok{return}\NormalTok{ registry}\OperatorTok{.}\FunctionTok{upsert}\NormalTok{(deploy)}\OperatorTok{;}
\NormalTok{\}}
\end{Highlighting}
\end{Shaded}

\texttt{entitlementGate.allows} is the narrow-default check. An
unconfigured business area falls back to a low ceiling.

\texttt{findSimilar} is the overlap gate. It embeds the new solution's
problem statement, retrieves the nearest existing registrations by that
embedding, and hands the pair to a small model that classifies how much
the two actually overlap. The two thresholds live on that
classification, not on raw cosine distance, because embedding similarity
alone is a poor proxy for whether two solutions do the same job. A low
bar stays purely informational: it's surfaced to reviewers but never
blocks. Only a high bar, a near-certain duplicate, returns a score above
\texttt{BLOCK\_THRESHOLD}.

\texttt{assessRisk} A step that reads what the solution actually is: the
instructions file, the resolved skill set (baked or live, Section 4.5),
and the tool groups it requests. It hands that to a larger model with
one job: decide whether this combination, for this business area, is
something a human should see before it ever runs. The same tool group
requested by a bot limited to read-only lookups reads differently from
that group requested alongside an instructions file that talks about
writing to a payment system. A larger reasoning model makes a call.

A failing registration is marked pending-review, neither rejected nor
silently allowed. CI/CD prints the reason into its own build summary and
exits cleanly. No infrastructure gets provisioned until the issue is
resolved. A review request posts to a team-chat channel with plain
approve/deny links. A reviewer clicking one hits the registry's own
decision endpoint directly, which flips both the review record and the
underlying registration status in a single step. The registry then
notifies the original requester by email and in the same chat channel,
with the outcome and, if approved, confirmation that re-running the
pipeline will now pass.

Figure 2 traces the CRM-triage solution from Section 4.7's configuration
example through these four mechanisms in the order it actually executes
at deploy time: config, registry, identity, gateway.

\begin{figure}
\centering
\pandocbounded{\includegraphics[keepaspectratio]{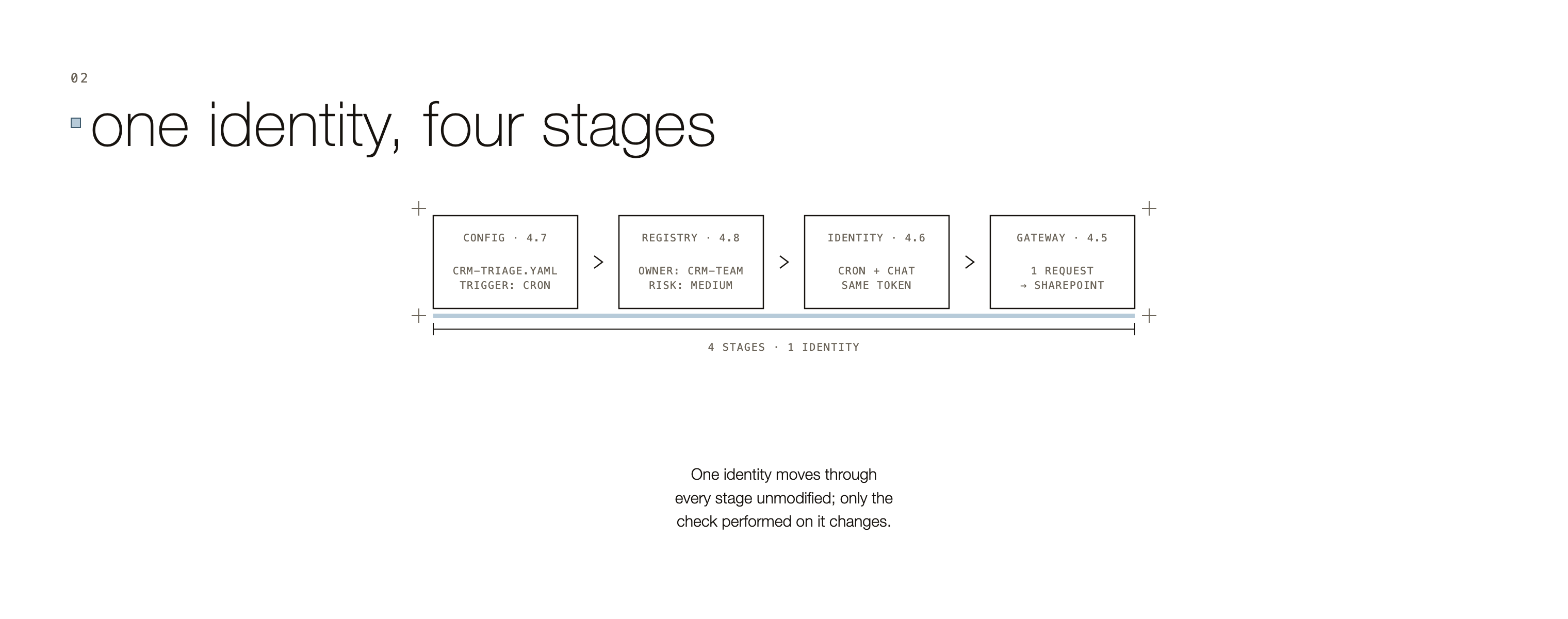}}
\caption{One identity moves through every stage unmodified; only the
check performed on it changes.}
\end{figure}

\subsection*{4.9 The business-user path}\label{the-business-user-path}
\addcontentsline{toc}{subsection}{4.9 The business-user path}

We assume a business user works with this architecture from a chat
surface already part of their daily work, such as Copilot Studio, Teams,
or Slack. From there, two separate entry points hit the same gateway and
the same identity model as the engineer's own path. Direct tool calls
let the chat assistant call a gateway-registered tool directly and
synchronously, which suits short, read-only lookups. A full harness run,
triggered as a chat action, addresses everything else. A multi-step
harness run will exceed a chat connector's timeout, so it can't be
called synchronously; ingress is split from execution into two separate
services instead. A small always-on endpoint validates the request,
hands it to a queue, and kicks off one execution of the actual solution
container. It then returns immediately:

\begin{Shaded}
\begin{Highlighting}[]
\KeywordTok{async} \KeywordTok{function} \FunctionTok{onChatAction}\NormalTok{(req}\OperatorTok{:}\NormalTok{ ChatActionRequest)}\OperatorTok{:} \BuiltInTok{Promise}\OperatorTok{\textless{}}\NormalTok{\{ accepted}\OperatorTok{:} \KeywordTok{true}\NormalTok{ \}}\OperatorTok{\textgreater{}}\NormalTok{ \{}
  \KeywordTok{const}\NormalTok{ caller }\OperatorTok{=} \ControlFlowTok{await} \FunctionTok{verifyCallerIdentity}\NormalTok{(req)}\OperatorTok{;}
  \ControlFlowTok{await}\NormalTok{ queue}\OperatorTok{.}\FunctionTok{enqueue}\NormalTok{(\{ solutionId}\OperatorTok{:}\NormalTok{ req}\OperatorTok{.}\AttributeTok{solutionId}\OperatorTok{,}\NormalTok{ caller}\OperatorTok{,}\NormalTok{ payload}\OperatorTok{:}\NormalTok{ req}\OperatorTok{.}\AttributeTok{payload}\NormalTok{ \})}\OperatorTok{;}
  \ControlFlowTok{return}\NormalTok{ \{ accepted}\OperatorTok{:} \KeywordTok{true}\NormalTok{ \}}\OperatorTok{;}
\NormalTok{\}}
\end{Highlighting}
\end{Shaded}

This is the entire run-trigger service. It holds no state past the
single \texttt{enqueue}. Execution itself runs as a separate, isolated
instance that picks the request back up, runs the full loop to
completion, and posts its own result to a callback once done. This split
exists for a concrete operational reason. The always-on endpoint is
subject to normal scale-in and redeploy behavior that would otherwise
kill a long-running loop mid-flight if it ran in the same process.

The identity carried through every hop of this chain is the end user's
own, not the calling service's. Tool visibility for a business user's
chat-triggered run is gated on that user's own access.

\subsection*{4.10 Code structure}\label{code-structure}
\addcontentsline{toc}{subsection}{4.10 Code structure}

Because Section 4.6 keeps every authorization decision outside the
harness process, the same harness codebase runs unmodified, under one
identity model, across all three deployment contexts described in
Sections 4.7 and 4.9: terminal, unattended cron backbone, and
chat-triggered execution. The architecture maps onto five repositories
or services. The harness itself is pulled and used locally, with
instructions (skills) and tools changed via a configuration file and
pushed.

The skill library is a git-versioned repository of shared procedural
documents described in Section 4.5. It's either live against an
engineer's own PAT while working interactively, or resolved per skill at
build time for a deployed solution: baked into the image by default, or,
for any skill a configuration names live, fetched at runtime by the
deployment's own scoped token (Section 4.7). The tool gateway and the
solution registry are each an always-on API. So is the run-trigger
service, which launches the harness's cron-deployed backend manually and
lets business users interact with it, for example via Copilot Studio.

\subsection*{4.11 Operating model}\label{operating-model}
\addcontentsline{toc}{subsection}{4.11 Operating model}

A small central platform team owns: the gateway, the lifecycle registry,
the reference harness repository, and the policy baseline. It doesn't
own use cases and doesn't build solutions for teams. Each business or
engineering unit owns its own use-case identification and delivery. A
small enablement function sits between the two, onboarding new tools to
the gateway, reviewing high-risk requests, running office hours. Its job
is to unblock a team by approving/rejecting. If a unit repeatedly asks
the enablement function to build for it, that's a signal the unit needs
different support. It doesn't mean the enablement function should grow
into a delivery team.

Registration replaces approval as the entry point. A team registers
first (owner, purpose, tools, business area); the deploy flow already
produces this as a side effect. It escalates to human review only when
the entitlement or overlap gate described above actually trips. A team
doesn't wait for a committee before it can start building. A recurring
sync between the central team and unit leads is the coordination point
that keeps the registry a living instrument: new registrations are
reviewed, infrastructure-tier graduations are discussed, and new tool
requests are triaged for the gateway.

Custom code used to justify an upfront gate because looking back at it
was expensive: reviewing a solution meant reading its code, and every
team's code looked different. Frontier models have already collapsed the
other half of that: an engineer can now build the exact custom thing
they saw a need for, cheaply. People embedded in their own domain see
the niche problems nobody else does. These problems are too small or too
specific for a central team to ever prioritize, but real enough to cost
someone hours every week. A central team that has to build everything
itself is stuck prioritizing by scale, and the niche problem never gets
solved. Letting the person who actually sees the problem build the fix
themselves, cheaply, is the whole point of coding getting cheap.

This architecture removes the cost that justified the gate. Every
registered solution runs on the same reference codebase, changed only
through its instructions file and configuration file (Section 4.10), so
reviewing what an enterprise's agent solutions actually do never
requires reading a codebase at all. Governance becomes the natural
outcome of the deployment path in Sections 4.7 and 4.8. When every
solution shares one identical codebase, auditing N solutions collapses
to reviewing N instruction files. Each is shorter and more legible than
the pull request it would otherwise have taken to ship the same change.

\subsection*{4.12 Durability of the
design}\label{durability-of-the-design}
\addcontentsline{toc}{subsection}{4.12 Durability of the design}

The architecture bets on continued frontier-model capability
improvement, a trend that has compounded for several years running. A
harness architecture converts that improvement directly into value for
every already-deployed solution, with zero code changes. A graph- or
chain-based architecture does not grant that same free upgrade, because
its behavior is frozen into whatever path the designer wired at build
time.

The bet is on capability improvement generally; no single lab has to be
the one supplying it. Open-weight models are closing the gap at a fast
clip and are a real hedge on the same durability argument. Nothing in
the proposed architecture ties it to one model provider.

\section*{5. Discussion / Limitations}\label{discussion-limitations}
\addcontentsline{toc}{section}{5. Discussion / Limitations}

No benchmark accompanies this paper. Every claim that harnesses suffice,
or outperform more elaborate architectures, rests on the work cited in
Section 2 plus field experience from the engagements described there. No
comparison was run for this paper itself.

The four failure modes named in arXiv:2604.13107 (lazy heuristics,
hallucination, dropped constraints, overconfidence) have no dedicated
mechanism addressing them in this paper. The deployment path in Section
4.7 remains the actual mitigation for all four. An engineer runs the
task live first, then distills what worked into the instructions file.
Experience across the engagements in Section 2 suggests this holds up.
We don't have a controlled benchmark showing how well this holds up, or
against which of the four modes specifically.

Governance by construction (Section 4.2) governs the path built on the
reference repository. It says nothing about the installed base described
in Section 4.1, the retrieval pipelines, custom chains, and hand-built
integrations teams were already running before this architecture
existed. It doesn't retroactively close whatever access those already
have. New work built this way is governed by construction; old work
isn't touched.

Credential-scoped tooling (Section 4.5) is only as good as the model's
ability to compose a correct call from documentation and general
knowledge. That holds well for widely used, well-documented surfaces
like Microsoft Graph. It is untested in this paper against internal,
undocumented, or unusually shaped enterprise APIs, where a hand-built
method might still beat a model composing its own request from scratch.

SHarD (arXiv:2607.25890, Section 2) names three controls for
distributing security through a harness: OS sandboxing, skill scanning,
tool restriction. This paper develops only the third in depth. Every
deployed solution already runs inside its own container (Section 4.7),
which provides real process- and namespace-level isolation. That's a
coarser guarantee than the dedicated OS sandboxing SHarD tests, things
like syscall filtering and scoped filesystem access within the sandbox
itself, and we haven't tested it against the boundary-crossing behavior
they report. Skill scanning has no equivalent in our paper. A baked-in
skill document is trusted content at build time (Section 4.5), and
nothing inspects it for injected instructions before it is baked into an
image. A skill named live (Section 4.5, 4.7) widens that gap rather than
narrowing it. It's trusted content at every run rather than once at
build. There's no rebuild step where a scan could be inserted, and no
image to diff against a previous known-good version.

The knowledge substrate (Section 4.4) inherits its ACL fidelity entirely
from the source system's own permission model. Microsoft Graph in
particular makes enumerating an item's effective permissions, as opposed
to its explicit ones, hard to get right. A mirror is only as governed as
the ACL groups it captured at sync time. It also trades staleness for
auditability. A document changed after the last sync cycle is invisible
to a run until the next one completes. That window is a deliberate cost
of the same mechanism that makes a run's knowledge reconstructible after
the fact, not an oversight to be tuned away.

Finally, the evidence base is European enterprise engagements across a
handful of sectors, evaluated specifically against Azure's identity and
policy primitives (Section 2). Whether the same logic ports to other
clouds and to on-premises settings is still an open claim. This paper
hasn't demonstrated it.

\section*{6. Conclusion}\label{conclusion}
\addcontentsline{toc}{section}{6. Conclusion}

This paper argued that the harness is a universally capable architecture
for enterprise AI automation, and gave the mechanism for running it that
way, governably, at enterprise scale. Four mechanisms do the work.
Credential-scoped tooling moves the curation to the model; not rigid
tool schemas. The same harness artifact runs unmodified across a
terminal, a business-facing chat surface, and an unattended cron
deployment; an enterprise no longer needs a separate system per surface.
Registration is a side effect of shipping and makes governance a natural
outcome of deployment. And because a harness can spawn other instances
of itself, a flagged call is judged by a fresh instance with no stake in
the task finishing, before a human is ever asked.

Each of these answers a gap the recent work in Section 2 names but does
not itself close. Real enterprise engagements informed the mechanisms in
this paper; none of it was designed in isolation.

This is a case for durability. It makes no claim to novelty. The
architecture bets on frontier-model capability continuing to compound.
Open-weight models make that bet easier to place, since nothing ties it
to one lab (Section 4.12). It also bets on governance staying cheap
enough that teams keep choosing the fast path voluntarily.

The gap this paper does not close is measurement. No benchmark
accompanies these claims. This follows the same move as Section 2's
disclosure argument. The paper discloses the reference implementation it
was built on and leaves the comparison against alternatives for later
work.

\section*{References}\label{references}
\addcontentsline{toc}{section}{References}

Bechard, P., Marquez Ayala, O., Chen, E., Skelton, J., Davasam, S.,
Sunkara, S., Yadav, V., Rajeswar, S. Terminal Agents Suffice for
Enterprise Automation. arXiv:2604.00073.
\url{https://arxiv.org/abs/2604.00073}

Ivanov, M., Rana, A., Prabhakaran, G. Can Coding Agents Be General
Agents? arXiv:2604.13107. \url{https://arxiv.org/abs/2604.13107}

Liu, J., Zhao, X., Shang, X., Shen, Z. Dive into Claude Code: The Design
Space of Today's and Future AI Agent Systems. arXiv:2604.14228.
\url{https://arxiv.org/abs/2604.14228}

Pan, K., Hou, R. Beyond Autonomy: A Dynamic Tiered AgentRunner Framework
for Governable and Resilient Enterprise AI Execution. arXiv:2605.10223.
\url{https://arxiv.org/abs/2605.10223}

Ning, X., Tieu, K., Fu, D., Wei, T., Li, Z., Bei, Y., Zou, J., Ai, M.,
Liu, Z., Li, T.-W., Chen, L., Zhao, Y., Yang, K., Li, B., Qian, C., Li,
G., Lin, X., Zeng, Z., Qiu, R., Chen, S., Sun, Y., Yang, X., Wang, R.,
Pan, R., Yang, C., Zhang, D., Fang, L., Cui, Z., Cao, Y., Chen, P., Sun,
D., Chen, R., Srinivasan, M., Mathur, N., Xia, Y., Li, H., Yan, H., Lu,
P., Zhang, L., Zhang, T., Tong, H., He, J. Code as Agent Harness.
arXiv:2605.18747. \url{https://arxiv.org/abs/2605.18747}

Zhang, Y., Wang, J., Ge, Y., Xu, W., Hamm, J., Reddy, C. K. Stop
Comparing LLM Agents Without Disclosing the Harness. arXiv:2605.23950.
\url{https://arxiv.org/abs/2605.23950}

Gore, W. R. Distributing Security Controls Through Harness Engineering.
arXiv:2607.25890. \url{https://arxiv.org/abs/2607.25890}

Anthropic. How We Built Our Multi-Agent Research System.
\url{https://www.anthropic.com/engineering/multi-agent-research-system}

micro-cc. Reference harness implementation.
\url{https://pypi.org/project/micro-cc/}

\end{document}